\documentclass[letterpaper, 10 pt, conference]{ieeeconf}  
\usepackage{silence}  
\IEEEoverridecommandlockouts                              

\usepackage{threeparttable}

\let\labelindent\relax
\usepackage{graphicx}
\usepackage{amsmath}
\usepackage{algorithmic}
\usepackage{booktabs} 
\usepackage{graphicx}
\usepackage{enumitem}

\usepackage{float}    
\usepackage{graphicx} 

\usepackage[ruled,vlined]{algorithm2e}

\usepackage{epstopdf}

\usepackage{graphics} 
\usepackage{epsfig} 
\usepackage{mathptmx} 
\usepackage{times} 
\usepackage{amsmath} 
\usepackage{amssymb}  
\usepackage{cuted} 
\usepackage{xcolor}

\usepackage[hypcap=false]{caption}
\usepackage{cite} 
\makeatletter  
\let\NAT@parse\undefined
\makeatletter

\usepackage[pagebackref,breaklinks, colorlinks=false, 
pdfborder={0 0 1},
citebordercolor={0 1 0},
linkbordercolor={1 0 0},
urlbordercolor={0 0 1}]{hyperref} 

\usepackage{glossaries}
\usepackage{multicol}

\usepackage{subcaption}
\usepackage{amsmath}
\title{%
  {%
    \vspace*{-2.6em}%
    \large\color{gray}%
    This paper has been accepted for publication at the\\[-0.5em]%
    Intelligent Vehicles Symposium (IV), Romania 2025. ©IEEE
  }%
  \\[1em]%
  {%
    \LARGE\bfseries
    Learning to Drift in Extreme Turning with Active Exploration and Gaussian Process Based MPC
  }%
}

\author{Guoqiang Wu\textsuperscript{*},  Cheng Hu\textsuperscript{*},  Wangjia Weng,
    Zhouheng Li,  Yonghao Fu, Lei Xie\textsuperscript{†}, and Hongye Su 
\thanks{This work was supported by the Ningbo Key research and development Plan (No.2023Z116).}%
\thanks{\textsuperscript{*}These authors contributed equally to this work.}
\thanks{\textsuperscript{†}The corresponding author of this paper.}
\thanks{Guoqiang Wu, Cheng Hu, Wangjia Weng, Zhouheng Li, Yonghao Fu, Lei Xie, and Hongye Su are with 
	the State Key Laboratory of Industrial, Zhejiang University, Hangzhou 310027, China.
{\tt\small \{22360409,22032081,22260379,zh.li,22360414\}
@zju.edu.cn; \{leix,hysu\}@iipc.zju.edu.cn}.}%
}

\begin{document}
\newacronym{gpr}{GPR}{Gaussian Process Regression}
\newacronym{aedgpr}{AEDGPR-MPC}{Active Exploration with Double-layer GPR and MPC}
\newacronym{ysc}{YSC}{Yaw Stability Control}
\newacronym{esc}{ESC}{Electronic Stability Control}
\newacronym{mpc}{MPC}{Model Predictive Control}
\newacronym{map}{MAP}{Model and Acceleration-based Pursuit }
\newacronym{pp}{PP}{Pure Pursuit}
\newacronym{lmpc}{LMPC}{Linear Model Predictive Control}
\newacronym{nmpc}{NMPC}{Nonlinear Model Predictive Control}
\newacronym{lqr}{LQR}{Linear Quadratic Regulator}
\newacronym{kmpc}{KMPC}{Koopman Model Predictive Control}
\newacronym{nn}{NN}{Neural networks}
\newacronym{rmse}{RMSE}{Root Mean Square Error}

\maketitle

\thispagestyle{empty}
\pagestyle{empty}



\begin{strip}
\vspace{-3.2cm}
\centering
\includegraphics[width=1\textwidth]{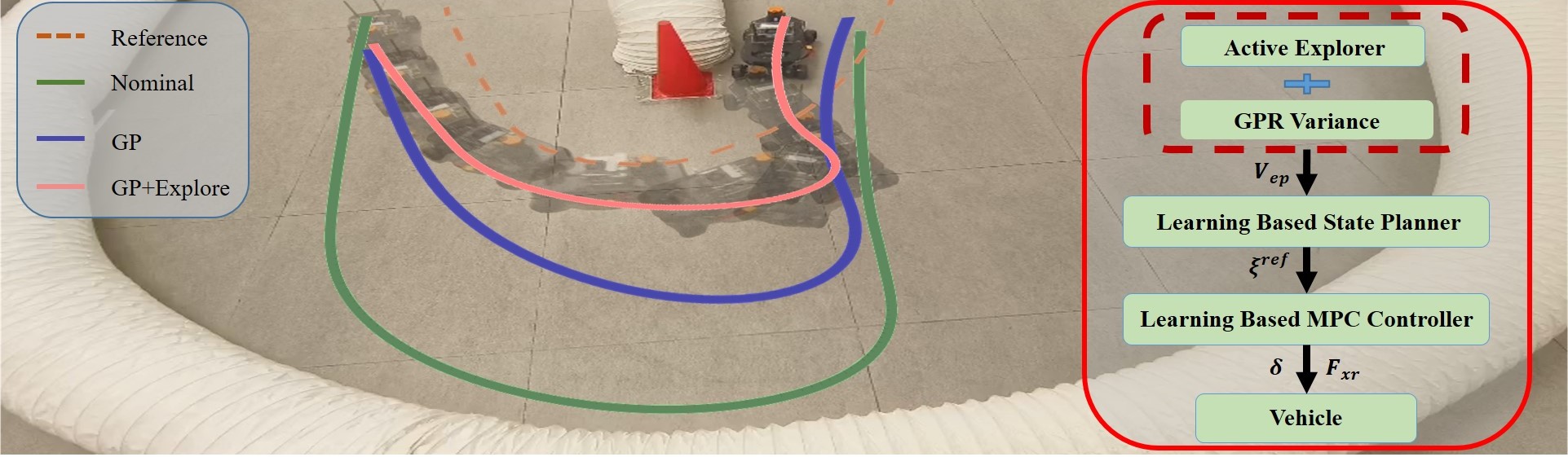}
\captionof{figure}{
  An aggressive drifting and cornering control framework with active exploration of \gls{gpr}.  
  \textbf{Step 1:} Employ the traditional \gls{mpc} drift controller for cornering and collect data.  
  \textbf{Step 2:} Fit a \gls{gpr} model to correct the vehicle model mismatch.  
  \textbf{Step 3:} Explore drifting behavior at various cornering speeds to enrich the \gls{gpr} dataset and determine the optimal cornering speed.
}

\label{fig: position drift of rc vehicle}
\vspace{-0.5cm}
\end{strip}


\begin{abstract}
Extreme cornering in racing often leads to large sideslip angles, presenting a significant challenge for vehicle control. Conventional vehicle controllers struggle to manage this scenario, necessitating the use of a drifting controller. However, the large sideslip angle in drift conditions introduces model mismatch, which in turn affects control precision. To address this issue, we propose a model correction drift controller that integrates Model Predictive Control (MPC) with Gaussian Process Regression (GPR). GPR is employed to correct vehicle model mismatches during both drift equilibrium solving and the MPC optimization process. Additionally, the variance from GPR is utilized to actively explore different cornering drifting velocities, aiming to minimize trajectory tracking errors. The proposed algorithm is validated through simulations on the Simulink-Carsim platform and experiments with a 1:10 scale RC vehicle. In the simulation, the average lateral error with GPR is reduced by 52.8\% compared to the non-GPR case. Incorporating exploration further decreases this error by 27.1\%. The velocity tracking Root Mean Square Error (RMSE) also decreases by 10.6\% with exploration. In the RC car experiment, the average lateral error with GPR is 36.7\% lower, and exploration further leads to a 29.0\% reduction. Moreover, the velocity tracking RMSE decreases by 7.2\% with the inclusion of exploration.

\end{abstract}


\section{INTRODUCTION}
Extreme turning scenarios in autonomous driving are particularly challenging and critical. Many control technologies have primarily focused on normal vehicle operating conditions, such as the \gls{pp} controller \cite{wang2017improved} and Model and Acceleration-based Pursuit\cite{10161472}. These systems limit the operating range of the tires to a linear zone in order to prevent tire slip and ensure a certain level of safety for the vehicle. However, in high sideslip angle cornering scenarios, these solutions struggle to maintain precise vehicle control.\par

You et al. demonstrated that a typical extreme cornering maneuver can be approximated by three distinct phases: entry corner guiding, steady-state sliding, and exit \cite{you2019high}. This suggests that extreme cornering can be effectively executed in a drifting state. Weng et al. further showed that the drifting state significantly enhances trajectory tracking accuracy during extreme cornering compared to regular state control \cite{weng2024aggressive}. By examining drifting techniques, we can ensure safe driving in high sideslip angle scenarios.\par


Most vehicle drifting controllers rely on the \gls{mpc} method. For \gls{mpc}-based drift controllers, two primary factors influence drift accuracy. The first factor is the calculation of the drift equilibrium point, which is essential for vehicle drift, as it directly affects the accuracy of the reference state. This equilibrium point depends on the vehicle's dynamic model, and any inaccuracies in the model can result in suboptimal performance. The second factor is the application of the \gls{mpc} controller for drift control, which uses the vehicle's dynamic model to optimize predictions over a rolling horizon. A mismatch between the vehicle model and the actual system can lead to a loss of control accuracy, further impairing drift performance. \par

To mitigate these issues, this paper employs the \gls{gpr} algorithm to compensate for vehicle model mismatches, ensuring both the accuracy of the drift equilibrium point calculation and the effectiveness of the \gls{mpc} control.\par

However, the performance of the \gls{gpr} algorithm is closely dependent on the richness of the collected data. To enhance the performance of \gls{gpr}, we conducted an exploration of the vehicle's cornering drift speed. This approach facilitates the acquisition of a more comprehensive dataset and helps determine the optimal drift cornering speed.

In summary, we propose a novel algorithm that reduces vehicle dynamics mismatch by leveraging \gls{gpr} and actively explores the optimal cornering drift speed. The primary contributions of this paper are as follows:

\begin{enumerate}[label=\Roman*] 

    \item \textbf{A double-layer \gls{gpr} is used to compensate for the vehicle model mismatch in the equilibrium point calculation and the \gls{mpc} controller.} Through the double compensation of the vehicle model, both the reference control state and \gls{mpc} control outcome are simultaneously improved.

     \item \textbf{A novel active exploration with \gls{gpr} method is developed.} A series of velocity sets with correct directions are generated. By leveraging the variance of \gls{gpr}, the algorithm actively explores regions with less information and identifies the optimal velocity. 
    \item \textbf{Experiments are conducted using a 1:10 scale RC car to validate the reliability of the algorithms.} In the RC car experiment, the average lateral error is reduced by 36.7\% in the \gls{gpr} compensated case compared to the non-\gls{gpr} case. This error is further reduced by 29.0\% through active exploration. The velocity tracking \gls{rmse} is also reduced by 7.2\% with active exploration. 

\end{enumerate}\par

\section{RELATED WORKS}
\subsection{Traditional Drift Controller}
Previous research on vehicle drift has primarily concentrated on developing various controllers based on complex vehicle dynamics models. The three-state single-track model, in particular, has been widely employed in dynamic surface, \gls{lqr}, and \gls{mpc} controllers \cite{park2021experimental, dong2022real, hu2024novel}. Additionally, a more detailed four-wheeled vehicle model has been used for steady-state drift analysis \cite{velenis2010stabilization}. However, the choice of vehicle model significantly influences controller performance. Accurately capturing the true dynamics of the vehicle remains a challenge for conventional methods.


\subsection{Learning-Based Model Predictive Control}
\gls{nn} are employed to mitigate inherent biases in vehicle dynamics models\cite{zhou2022learning,9772321}. However, evaluating the effects of learning remains challenging due to the complex architectures of neural networks. The Koopman operator is a tool that maps nonlinear systems to a high-dimensional space and constructs linear models to describe the nonlinear evolution of the system \cite{ joglekar2023analytical, joglekar2023data}. Nonetheless, adapting the Koopman operator to more complex nonlinear vehicle dynamics models is difficult, and ensuring control stability presents a significant challenge. Kabzan et al. incorporate \gls{gpr} to account for residual model uncertainty, thereby achieving safe driving behavior in racing scenarios \cite{kabzan2019learning}. Su et al. employ two \gls{gpr}-based error compensation functions to adjust both the planner's model and the controller's model, resulting in improved racing performance \cite{su2023double}. \gls{gpr} enhances model accuracy and evaluates the learning effect by analyzing variance. The diversity of datasets is crucial for the performance of \gls{gpr}. Thus, exploring diverse datasets is vital for improving \gls{gpr} capabilities \cite{schreiter2015safe}. However, the aforementioned studies do not address dataset exploration in the context of vehicle drifting.

\begin{table}[htbp]
\caption{Comparison with related innovative methods.}
\scriptsize
\renewcommand{\arraystretch}{1.5}
\centering
\begin{threeparttable}
\begin{tabular}{@{}l|l|c|l@{}}
\toprule
\textbf{Method} & \textbf{Model} & \textbf{Drift ability}{*} & \textbf{Learning ability\textsuperscript{†}} \\  
\midrule
\textbf{\gls{pp}}\cite{wang2017improved} & None & Incapable & Incapable  \\
\textbf{\gls{lqr}}\cite{park2021experimental} & Requires & Capable & Incapable  \\
\textbf{\gls{mpc}}\cite{hu2024novel} & Requires & Capable & Incapable  \\
\textbf{\gls{gpr}-\gls{mpc}}\cite{su2023double} & Requires & Incapable & Capable  \\
\textbf{ADEGPR-MPC(ours)} & Requires & Capable & Capable  \\

\bottomrule
\end{tabular}
\begin{tablenotes}
\footnotesize
\item[*] \textbf{Drift ability:} Capable means related works having the ability to drift.
\item[†] \textbf{Learning ability:} Capable means related works having the ability to learn the vehicle model from data.
\end{tablenotes}
\label{tab:control_methods_transposed}
\end{threeparttable}
\end{table}

Section III of this research presents our nominal vehicle model and related knowledge on \gls{gpr}. The \gls{aedgpr} structure is explained in section IV. The simulation result on the Matlab-Carsim platform is shown in Section V. In Section VI, experiments using a 1:10 scale RC car are presented. The conclusion of the paper is presented in Section VII.

\section{Learning Vehicle Model}
This section introduces the nominal vehicle dynamics model, which serves as the foundation for the drifting controller but lacks precision. Subsequently, we present the theory of \gls{gpr}, which compensates for the biases inherent in the nominal model.\par

\subsection{Nominal vehicle model}
\begin{figure}[thpb]
    \centering
    \includegraphics[width=0.35\textwidth]{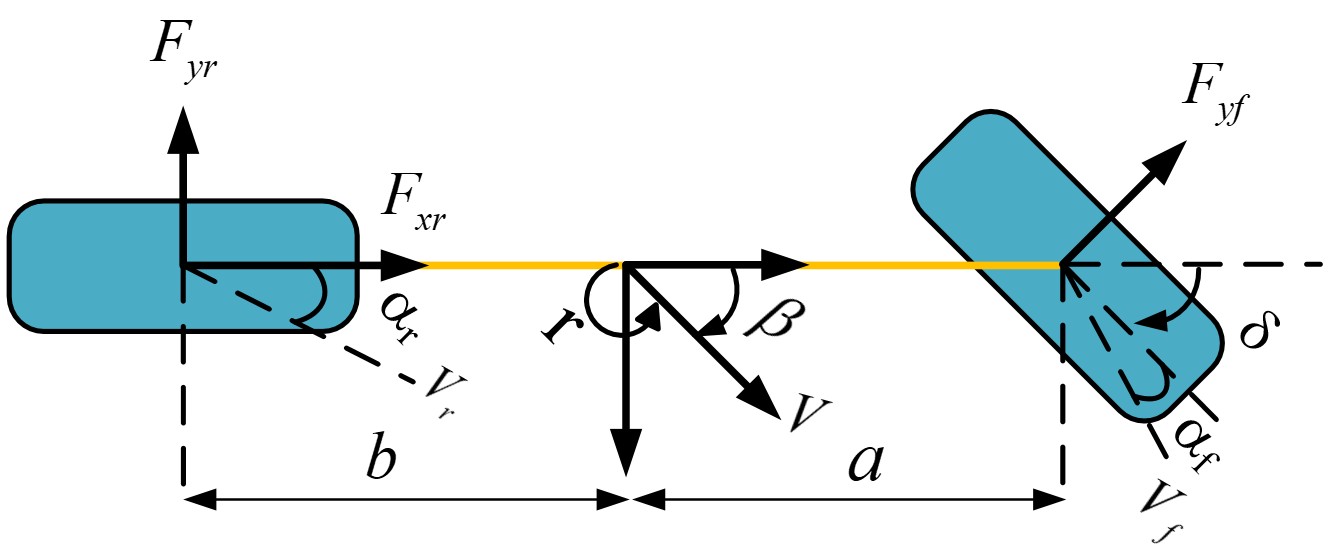}
    \caption{Single-track vehicle dynamics model.}
    \label{fig:Vehicle_model}
\end{figure}

One of the most widely used vehicle dynamics models is the single-track model, which is commonly employed in drift controllers. This model provides an effective representation of a real vehicle using a limited set of fundamental parameters. As illustrated in Fig.~\ref{fig:Vehicle_model}, we adopt this model as the nominal vehicle model in our approach.

The equations for this vehicle model can be expressed as follows:
\begin{equation}
\begin{aligned}
\dot{V} &= \frac{-F_{yf} \sin(\delta - \beta) + F_{yr} \sin \beta + F_{xr} \cos \beta}{m}, \\
\dot{\beta} &= \frac{F_{yf} \cos(\delta - \beta) + F_{yr} \cos \beta - F_{xr} \sin \beta}{mV} - r, \\
\dot{r} &= \frac{a F_{yf} \cos \delta - b F_{yr}}{I_z},
\end{aligned}
\label{eq:nominal_model}
\end{equation}
where \( r \) denotes the yaw angular velocity, \( \beta \) is the sideslip angle, \( \delta \) represents the angle of the front wheel, \( m \) is the vehicle mass, and \( I_z \) is the moment of inertia. Furthermore, \( V \) indicates the velocity of the vehicle's center of mass, \( F_{yf} \) and \( F_{yr} \) are the lateral forces acting on the front and rear wheels, \( F_{xr} \) is the longitudinal force on the rear wheels, and \( a \) and \( b \) represent the distances from the center of gravity to the front and rear wheels, respectively. 
\par
The complexity of vehicle models is significantly influenced by tire models, particularly when driving at high velocities and during drifting. In this paper, we implement the simplified Pacejka Tire Model \cite{bakker1987tyre} as follows:
\begin{equation}
\begin{aligned}
F_y=-\boldsymbol{\mu}F_z\sin\left(C\arctan\left(B\boldsymbol{\alpha}\right)\right),
\end{aligned}
\label{eq: Pacejka Tire}
\end{equation}
where \( F_y \) denotes the lateral force acting on the front or rear tire, \( F_z \) represents the vertical load on the tires, \( B \) and \( C \) are tire parameters that need to be identified, \( \mu \) is the coefficient of friction between the tire and the ground, and \( \alpha \) is the tire slip angle. The formulas for calculating the slip angles of the front and rear tires are as follows:
\begin{equation}
\begin{aligned}
\boldsymbol{\alpha}_{f}&=\arctan{(\frac{V\sin\boldsymbol{\beta}+ar}{V\cos\boldsymbol{\beta}})}-\boldsymbol{\delta},\\
\boldsymbol{\alpha}_{r}&=\arctan{(\frac{V\sin\boldsymbol{\beta}-br}{V\cos\boldsymbol{\beta}})}.
\end{aligned}
\label{eq: Tire_slip_angle}
\end{equation}

\subsection{Learning-Based Vehicle Model}
Although the nominal vehicle dynamics model is sufficient for the sample drifting controller, our goal is to enhance drifting performance by learning the vehicle model error \( \textbf{d} \) using \gls{gpr}, a non-parametric learning method.

The compensated vehicle model with \gls{gpr} can be formulated as follows:
\begin{equation}
\begin{aligned}
\mathbf{\dot{x}}=\mathbf{f(x, u)}+\mathbf{d(x, u)},
\end{aligned}
\label{eq: real_error_model}
\end{equation}
where \( \mathbf{x} \) represents the system state \( [V, \beta, r]^{T} \), and \( \mathbf{u} \) denotes the system inputs \( [\delta, F_{xr}]^T \). \( \textbf{f} \) refers to the nominal model in equation \eqref{eq:nominal_model}, and \( \mathbf{d} \) represents the model error between the real vehicle model and the nominal model. The \gls{gpr} is employed to fit the model error \( \mathbf{d}\in {\mathbb{R}^{3}} \).\par

The following provides an explanation of the principles related to \gls{gpr}. Let \( \mathbf{z} \in \mathbb{R}^{n_f} \) represent the feature vector, where \( n_f \) is the number of dimensions, and \( \mathbf{d} \in \mathbb{R}^{n_d} \) represent the output vector corresponding to the model error, where \( n_d \) is the number of dimensions. We assume that they are related as follows:

\begin{equation}
\begin{aligned}
\mathrm{\mathbf{d(z)}\sim\mathcal{N}(\mathbf{\mu(z)}, \Sigma(\mathbf{z}))},
\label{eq: d_error_gp1}
\end{aligned}
\end{equation}
where $\mathbf{\mu(z)} = [\mu^1(\mathbf{z}), . . . , \mu^{n_d} (\mathbf{z})] \in \mathbb{R}^{n_d} $ and $\Sigma(\mathbf{z}) =diag([\Sigma^1(\mathbf{z}), . . . , \Sigma^{n_d}(\mathbf{z})]) \in \mathbb{R}^{n_d \times n_d}$. \par

Given a finite dataset of size $m$ consisting of feature output tuples, $\{(\mathbf{z}_1, \mathbf{d}_1), . . . , (\mathbf{z}_m, \mathbf{d}_m)\}$, we denote it as $\Lambda  =\{\mathbf{Z}, \mathbf{D}\}$ with input features $\mathbf{Z} = [\mathbf{z}^{T}_{1} ; . . . ; \mathbf{z}^{T}_{m}] \in \mathbb{R}^{m \times n_f}$ , and output data ${\mathbf{D} = [\mathbf{d}^{T}_{1} ; . . . ; \mathbf{d}^{T}_{m}]} \in {\mathbb{R}^{m \times n_d}}$, \gls{gpr} use the dataset $\Lambda$ to fit $\mu(\mathbf{z})$ and $\Sigma(\mathbf{z})$ as:

\begin{equation}
\begin{aligned}
&\mu^{a}(\mathbf{z})=\mathbf{k}_{\mathbf{zZ}}^{a}(\mathbf{K}_{\mathbf{ZZ}}^{a}+\mathbf{I}\sigma_{a}^{2})^{-1}[\mathbf{D}]_{.,a},\\
&\Sigma^{a}(\mathbf{z})=k_{\mathbf{zz}}^{a}-\mathbf{k}_{\mathbf{zZ}}^{a}(\mathbf{K}_{\mathbf{ZZ}}^{a}+\mathbf{I}\sigma_{a}^{2})^{-1}\mathbf{k}_{\mathbf{Zz}}^{a}.
\label{eq: d_error_gp2}
\end{aligned}
\end{equation}
For \(a = 1, \dots, n_d\), \(\sigma_a^2\) represents Gaussian noise, \([\mathbf{D}]_{.,a}\) denotes the \(a\)-th column of matrix \(\mathbf{D}\), and \(\mathbf{K}^a_{\mathbf{ZZ}}\) is the Gram matrix, where \([\mathbf{K}^a_{\mathbf{ZZ}}]_{ij} = k^a(\mathbf{z}_i, \mathbf{z}_j)\). Additionally, \([\mathbf{k}_{\mathbf{Zz}}^a]_j = k^a(\mathbf{z}_j, \mathbf{z}) \in \mathbb{R}\), and \(\mathbf{k}_{\mathbf{Zz}}^a = (\mathbf{k}_{\mathbf{zZ}}^a)^T \in \mathbb{R}^m\), and \(k_{\mathbf{zz}}^a = k^a(\mathbf{z}, \mathbf{z}) \in \mathbb{R}\), where \(k^a\) is the kernel function. For further details, readers are referred to \cite{wang2023intuitive} and \cite{schulz2018tutorial}.

\section{\glsentrytext{aedgpr} SYSTEM}

\begin{figure*}[thpb]
   \vspace{10pt}
    \centering
    \includegraphics[width=0.8\textwidth]{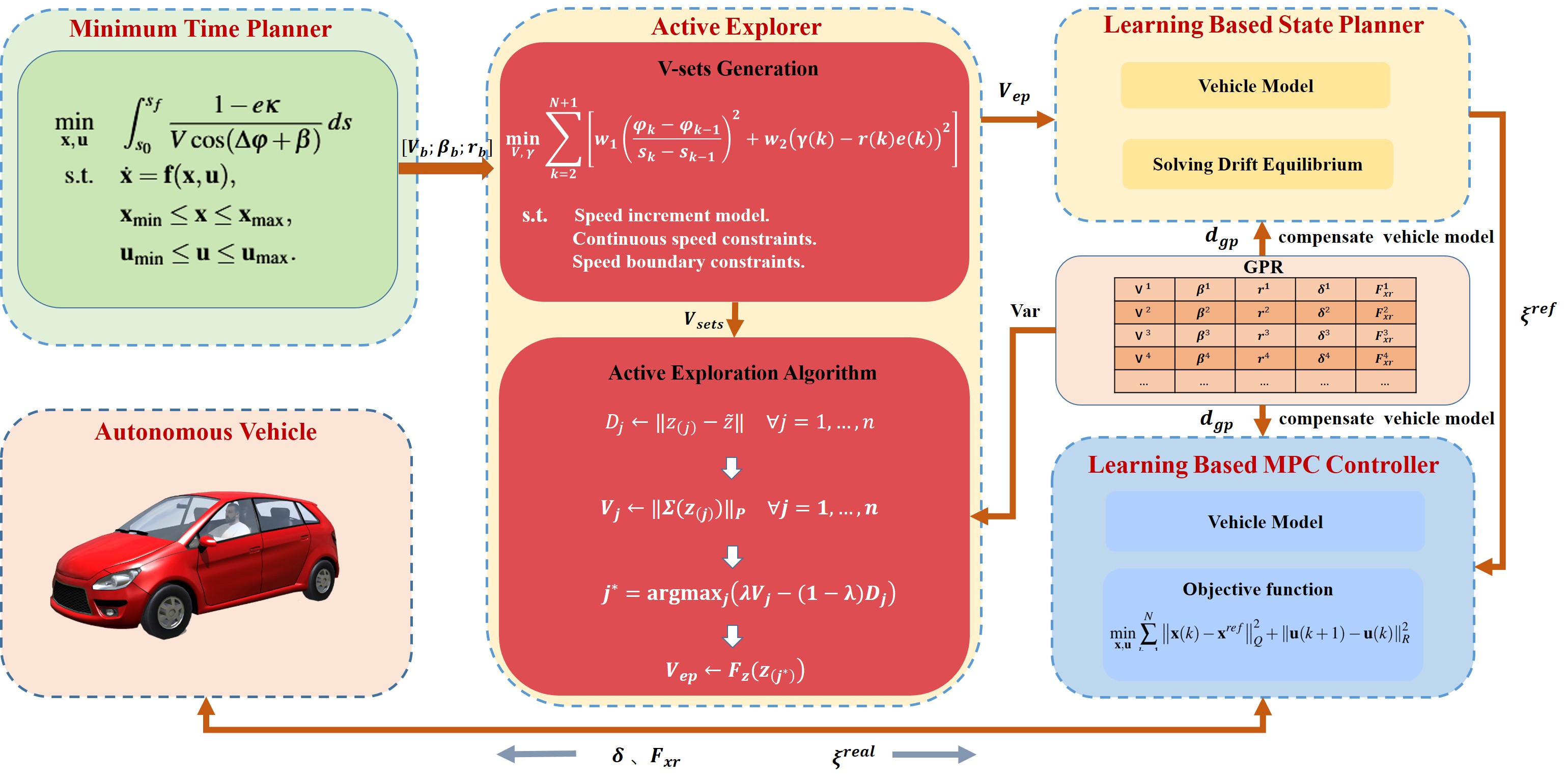}
   \caption{
  \gls{aedgpr} system architecture for extreme turning. The system consists of four modules: 
  \textbf{(1)} Minimum Time Planner module generates the reference trajectory and state. 
  \textbf{(2)} Active Explorer module generates candidate speeds for cornering and selects a speed \( V_{ep} \) for exploration. 
  \textbf{(3)} Learning-Based State Planner utilizes a mismatch correction model with \gls{gpr} to plan cornering reference states with \( V_{ep} \). 
  \textbf{(4)} Learning-Based \gls{mpc} Controller uses the same \gls{gpr}-corrected model to track cornering reference states and perform the drift maneuver.
}

\label{fig:overview}
\end{figure*}

The \gls{aedgpr} system is designed to address the challenge of extreme cornering using the drift technique. The structure of the \gls{aedgpr} system is shown in Fig.~\ref{fig:overview} and consists of four main components. The minimum-time planner module generates the reference trajectory and state for the active explorer module. Upon receiving the reference, the active explorer module first generates a set of candidate speeds for cornering and selects the speed \( V_{ep} \) to explore using the active exploration algorithm. The selected \( V_{ep} \) is then passed to the learning-based state planner module, which plans the corresponding cornering reference states by utilizing a mismatch correction model with \gls{gpr}. Finally, the learning-based \gls{mpc} controller, using the same \gls{gpr} model, tracks the cornering reference states to execute the drift maneuver.

\subsection{Minimum Time Planner}
The minimum-time planner is essentially an optimization problem based on the vehicle model. Before introducing it, we first discuss the concept of curvilinear coordinates. Curvilinear coordinates are employed to better represent the relationship between the vehicle and the reference trajectory. The formula is defined as follows:

\begin{equation}
\begin{aligned}
\dot{s} &= \frac{V\cos(\Delta\varphi+\beta)}{1-e\kappa}, \\
\dot{e} &= V\sin(\Delta\varphi+\beta), \\
\dot{\theta} &= \dot{\varphi}-\dot{\varphi}_{ref}, 
\label{eq: Curvilinear Coordinate system}
\end{aligned}
\end{equation}
where \( s \) represents the distance along the reference trajectory of the roadway, \( \varphi \) denotes the heading angle, \( \beta \) is the vehicle sideslip angle, and \( e \) represents the distance between the vehicle's center of gravity and the reference trajectory. Let \( \theta \) denote the difference between the vehicle's heading angle and the reference track angle. The parameter \( \kappa \) characterizes the curvature of the reference trajectory.\par
To achieve minimum-time state planning, we integrate the vehicle dynamics model \eqref{eq:nominal_model} with the curvilinear coordinate system \eqref{eq: Curvilinear Coordinate system} to derive the optimization problem \eqref{eq: min time planner}. The system state is denoted as \( \mathbf{x} = [ V, \beta, r, s, e, \theta ]^T \), and the system control input is represented as \( \mathbf{u} = [ \delta, F_{xr} ]^T \).\par

\begin{equation}
\begin{alignedat}{2}
\min_{\mathbf{x},\, \mathbf{u}} \quad & \int_{s_0}^{s_f} \frac{1 - e\kappa}{V \cos(\Delta\varphi + \beta)} \, ds \\
\text{s.t.} \quad 
& \dot{\mathbf{x}} = \mathbf{f}(\mathbf{x}, \mathbf{u}), \\
& \mathbf{x}_{\text{min}} \leq \mathbf{x} \leq \mathbf{x}_{\text{max}}, \\
& \mathbf{u}_{\text{min}} \leq \mathbf{u} \leq \mathbf{u}_{\text{max}}.
\end{alignedat}
\label{eq: min time planner}
\end{equation}

The optimization aims to minimize total travel time by minimizing the integral of inverse longitudinal speed, thus planning a minimum-time trajectory. By solving this optimization problem, the optimal planning state corresponding to the vehicle's minimum time can be obtained, denoted as \( [V_b, \beta_b, r_b]^T \).\par

Since equation (\ref{eq: min time planner}) represents a nonlinear optimization problem, its solution may not guarantee optimal control performance. Therefore, an exploration mechanism for cornering drift speed is introduced. The planned velocity \( V_b \) will serve as the baseline velocity for the active explorer module. The specific mechanism of the active explorer is described in the next section.\par

\begin{figure}[htbp]
    \centering
    \includegraphics[width=0.9\linewidth]{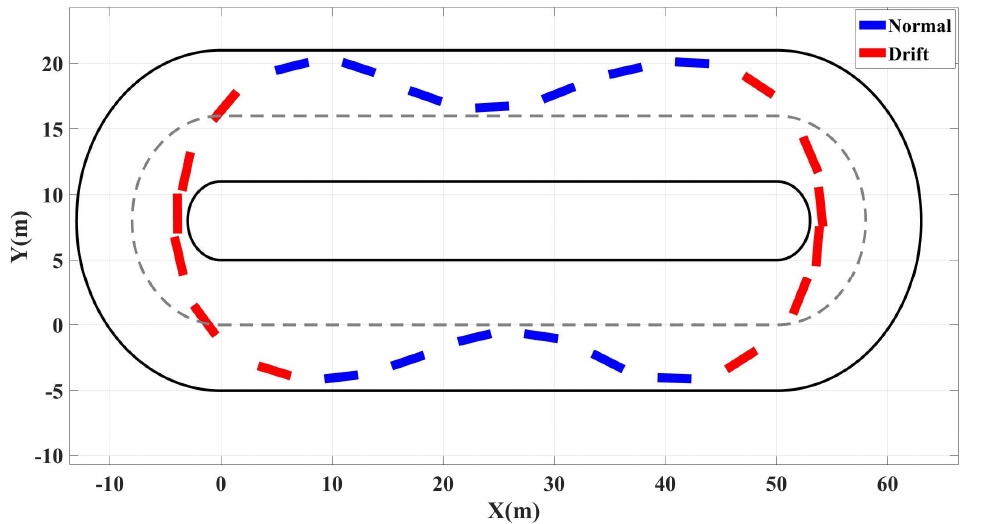}
    \caption{
    Planned minimum time trajectory in the simulation. The blue segments represent the vehicle operating in a normal state, while the red segments correspond to the drift state.
}
    \label{fig:plan_traj}
\end{figure}

\begin{figure}[htbp]
    \centering
    \includegraphics[width=0.45\textwidth]{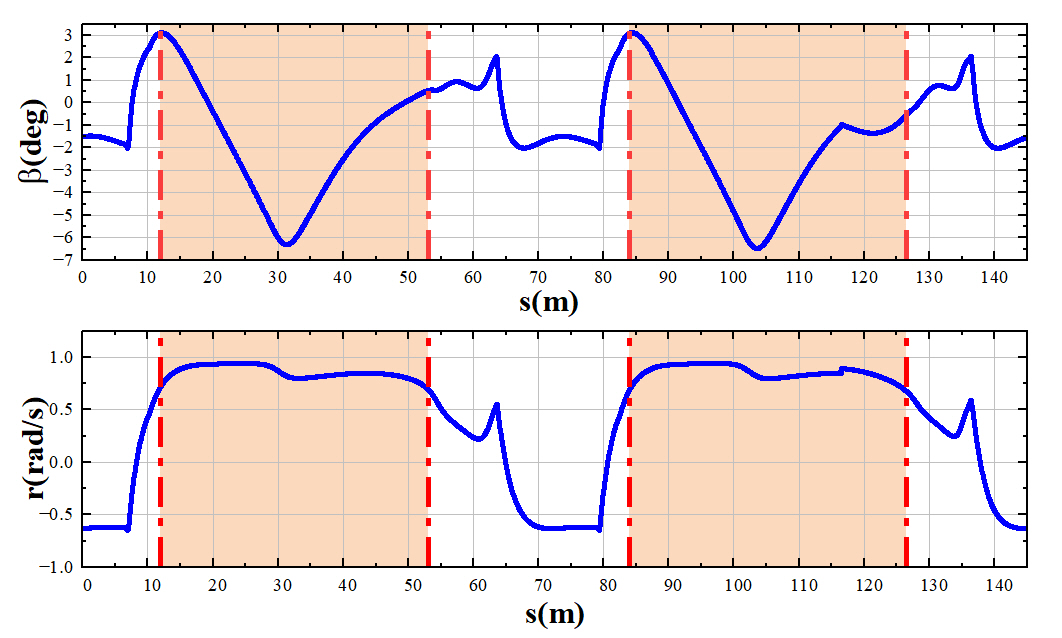}
    \caption{
    Sideslip angle and yaw rate of the minimum time planner in the simulation. 
    The orange section represents the car's cornering area, where the opposing directions of the sideslip angle \( \beta \) and yaw rate \( r \) indicate a drifting condition.
}

    \label{fig: planner state}
\end{figure}
Given a track for the playground, the minimum-time trajectory can be obtained by solving the corresponding optimization problem, as shown in Fig. \ref{fig:plan_traj}. The blue segments represent the vehicle operating in a normal state, while the red segments correspond to the drifting state. The planned minimum-time results for the sideslip angle \( \beta \) and yaw rate \( r \) are depicted in Fig.\ref{fig: planner state}. The orange section represents the car's cornering area, where it is evident that the sideslip angle \( \beta \) and yaw rate \( r \) are in opposing directions, indicating a drifting condition. Different controllers are then employed based on the vehicle's state: a \gls{pp} controller is used for non-drift states, while the \gls{aedgpr} system is employed for drift states.

\subsection{Active Explorer}
The active explorer module has two main functions. The first is to generate sets of speeds \( V_{\text{sets}} \) that can be explored. The second is to balance each vehicle state within the exploration sets and carefully select the next speed \( V_{ep} \) to explore, continuing this process until all sets have been explored.

\subsubsection{\textbf{V-sets Generate}}
Regarding the generation of the speed sets \( V_{\text{sets}} \), the following equation is provided:
\begin{equation}
\begin{aligned}
\min_{V, \gamma} \ & \sum_{k=2}^{N+1} \left[w_{1}\left(\frac{\varphi_{k}-\varphi_{k-1}}{s_{k}-s_{k-1}}\right)^{2} + w_{2}\left(\gamma(k) - r_b(k)e(k)\right)^{2}\right],  \\
\text{s.t.} \quad \ & V^{\text{new}}_{k} = V^{\text{origin}}_{k} + \gamma(k), \\
& \varphi_{k} = \arctan\left(\frac{\Delta V^{\text{origin}}_{k} + \Delta \gamma(k)}{\Delta s(k)}\right), \\
& V^{\text{new}}_{k} \in \begin{bmatrix}V_{k,\min}, V_{k,\max}\end{bmatrix}, \quad \forall \, 1 \leq k \leq N, \\
& \gamma(k) \in \begin{bmatrix}\gamma_{k,\min}, \gamma_{k,\max}\end{bmatrix}, \quad \forall \, 1 \leq k \leq N,
\end{aligned}
\hspace{-0.5cm}
\label{eq:Vsets gen}
\end{equation}
where \( w_1 \) and \( w_2 \) are the weighting parameters. The first term of the objective function is designed to ensure the smoothness of the velocity, where \( \varphi_k \) represents the curvature of the generated velocity curve and \( s(k) \) represents the distance along the roadway's reference trajectory. The second term, \( e(k) \), is derived from the lateral error between the actual vehicle trajectory and the planned minimum-time trajectory, ensuring that the exploration proceeds in the correct direction. \( r_b(k) \) indicates the yaw rate obtained from the minimum-time planner module. \( \gamma(k) \) denotes the increment used to generate a new velocity based on the original velocity. \( V^{\text{origin}}_k \) is the initially planned cornering speed point of the vehicle by the minimum-time planner, and \( V^{\text{new}}_k \) denotes the generated speed point that requires exploration.\par

For vehicle drift, the drift radius can be calculated using the formula \( R = \frac{V}{r} \). Therefore, by determining the lateral distance \( e(k) \) to either the inner or outer side of the reference trajectory, a set of velocities can be generated that ensures the correct direction for trajectory correction.\par

\begin{figure}[thpb]
    \centering
    \vspace{6pt}
    \includegraphics[width=0.42\textwidth]{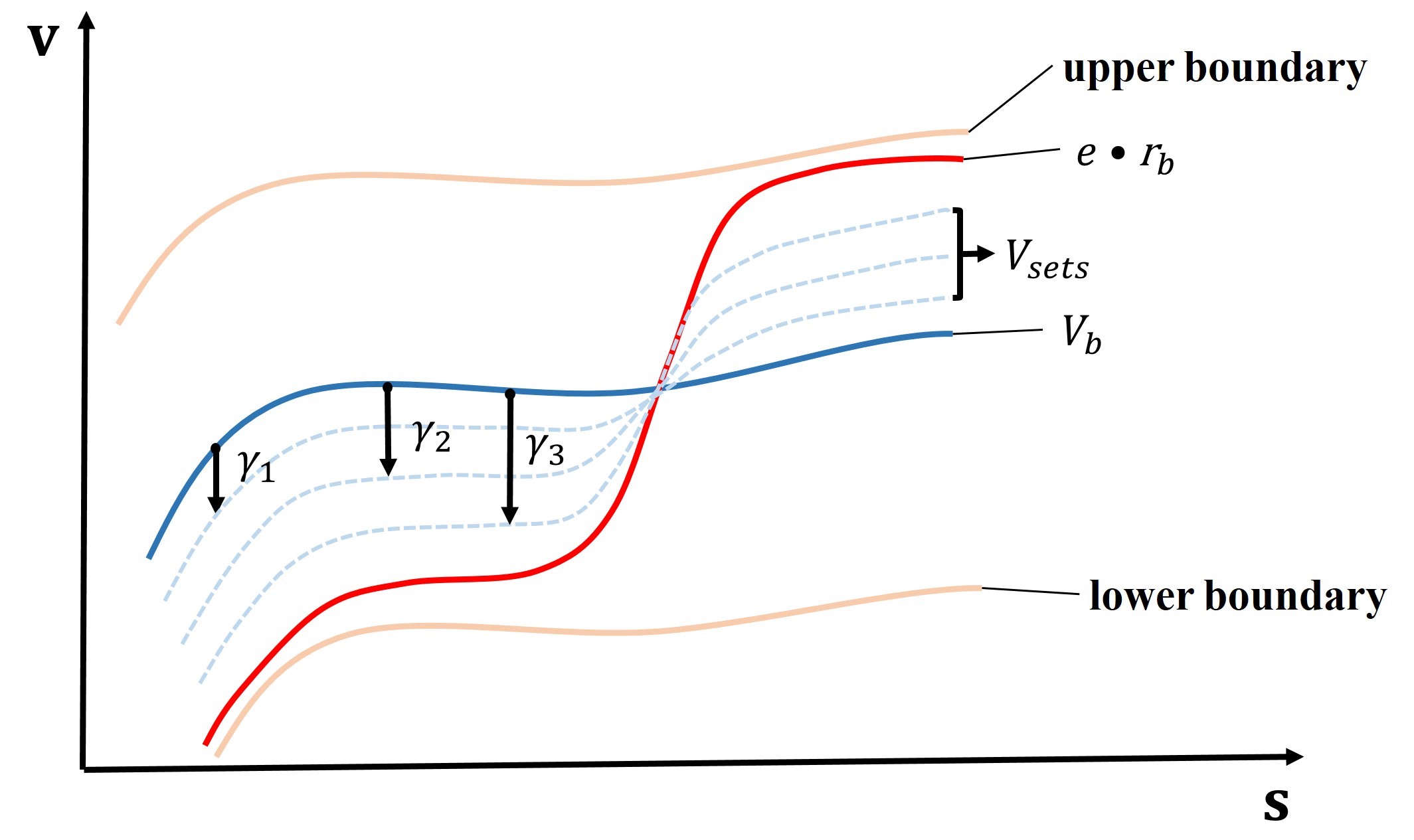}
     \caption{
    Schematic for the building of velocity sets. The blue solid line represents the initial vehicle speed, the red solid line shows the trend of generated speed sets calculated by $ e \cdot r_b $, and the blue dashed line indicates the generated speed series $ V_{\text{sets}} $.
}
    \label{fig:Velocity set generation process}
   
\end{figure}

The process of generating velocity sets is illustrated in Fig. \ref{fig:Velocity set generation process}. The orange solid line represents the upper and lower boundary constraints for the generated vehicle speeds. According to the first term of the objective function in the optimization problem (\ref{eq:Vsets gen}), the generated speed increment is expected to approach \( e \cdot r_b \) (the red solid line in the figure). However, the increment is constrained by the boundaries \( [\gamma_{\min}, \gamma_{\max}] \), which limit the speed increment \( \gamma \). By adjusting these increment boundaries, a series of valid speed increments \( \gamma \) can be generated, thereby forming the velocity sets \( V_{\text{sets}} \).

When the sets of speed \( V_{\text{sets}} \) have been generated, they will be used to calculate \( n \) candidate sets \( \mathbf{z}_j \) for \( j = 1, \dots, n \) by solving the drift equilibrium. The set \( \mathbf{z}_j \) represents the \( j \)-th candidate set. The input to the \gls{gpr} model comprises five components: speed \( V \), sideslip angle \( \beta \), yaw rate \( r \), front-wheel angle \( \delta \), and rear-wheel drive force \( F_{xr} \), denoted as \( \mathbf{z} = [V, \beta, r, \delta, F_{xr}]^T \).
\subsubsection{\textbf{Active Exploration Algorithm}} The algorithm is presented in Algorithm \ref{Active exploration algo}. The first step is to calculate the \gls{gpr} input \( \tilde{\textbf{z}} \) from the reference state \( \mathbf{\xi}^{\text{ref}} \), which the vehicle follows. Next, the distance \( D_j \) between the reference \gls{gpr} input \( \tilde{\textbf{z}} \) and the candidate \( \mathbf{z}_j \) is computed. Furthermore, we fully utilize the uncertainty estimation of \gls{gpr}, along with the weight matrix \( P \), to compute the variance \( V_j \) of each candidate set. Moreover, we balance the trade-off between distance and variance using the weight parameter \( \lambda \in [0,1] \), and select the maximizing feature \( \textbf{z}^{\text{ref}} \). Once the maximizing feature is selected, the velocity \( V_{ep} \) required for exploration is extracted from the maximizing feature \( \mathbf{z}^{\text{ref}} \). The reference is then updated by solving the drift equilibrium: \( \mathbf{\xi}^{\text{ref}} \gets E(V_{ep}, r_b) \). Subsequently, the vehicle is directed to follow the updated reference state and gather \gls{gpr} data. This process continues iterating until all speed sets have been explored.

\begin{algorithm}[htbp]
\caption{Active exploration via \gls{gpr} variance and distance}
\SetAlgoLined
\KwData{$\mathbf{\xi}^{\text{ref}}$, $\{ \mathbf{z} \}_n$, $\Sigma(\cdot)$, ${P}$, $\lambda$}
\KwResult{Best drift state $\mathbf{\xi}^{\text{ref}}$ and enriched \gls{gpr}}

\While{Not finished exploring}{
    \textbf{1) Get the \gls{gpr} input vector from reference:}
    
    \quad $\mathbf{\tilde{z}} \gets \phi_z(\mathbf{\xi}^{\text{ref}})$ 
    
    \textbf{2) Calculate candidate distance to reference:}
    
    \quad $D_j \gets {\|{\mathbf{z}}_{(j)} - \mathbf{\tilde{z}}\|} \quad \forall j = 1,\dots, n$
    
    \textbf{3) Solve candidate variance:}
    
    \quad $V_j \gets \|\Sigma({\mathbf{z}}_{(j)})\|_{{P}} \quad \forall j = 1,\dots, n$
    
    \textbf{4) Trade-off distance and variance:}
    
    \quad  ${\mathbf{z}^{\text{ref}} \gets \mathbf{z}_{(j^*)}}$, ${j^* = \arg\max_j \left(\lambda V_j - (1 - \lambda) D_j\right)}$
    
    \textbf{5) Extract the speed of exploration:}
    
    \quad $V_{\text{ep}} \gets F_z(\mathbf{z}^{\text{ref}})$

    \textbf{7) Update reference using equilibrium:}
    
    \quad $\mathbf{\xi}^{\text{ref}} \gets E(V_{ep},r_{b})$
    
    \textbf{6) Vehicle running and \gls{gpr} data collection}
}

\label{Active exploration algo}

\end{algorithm}

The key insight of this algorithm is its ability to leverage the variance of the \gls{gpr} to explore information-scarce regions and enrich the dataset. While \gls{gpr} struggles in data-limited areas, the algorithm ensures stable control by gradually exploring through the balance of distance and variance. In the cornering drift scenario, the \gls{gpr} dataset is enhanced through exploration, ultimately identifying the optimal cornering drift state.\par

\subsection{Learning-Based State Planner}
The learning-based state planner determines the cornering states by solving for drift equilibrium points. This section utilizes the learned vehicle dynamics model with \gls{gpr} to derive these equilibrium points. We consider five variables, denoted as \( [V, \beta, r, \delta, F_{xr}]^T \). We choose to fix the variables \( [V, r]^T \) and solve for the remaining variables under the conditions \( \dot{V} = \dot{\beta} = \dot{r} = 0 \). The resulting drift equilibrium points, denoted as \( \mathbf{\xi}^{\text{ref}} = [V_{ep}, \beta_{eq}, r_b]^T \), are then transmitted to the controller as the reference vehicle state. It is important to note that the fixed speed \( V \) is derived from the active explorer module, \( V_{ep} \), which indicates the areas requiring exploration, while \( r \) is obtained from the minimum time planner, denoted as \( r_b \).

\subsection{Learning-Based \glsentrytext{mpc} Controller}
We propose a learning-based \gls{mpc} controller that utilizes \gls{gpr} to learn the deviation between the actual model and the nominal model, leading to more precise control. The \gls{mpc} cost function is defined as follows:

\begin{equation}
\begin{aligned}
\min_{\mathbf{x}, \mathbf{u}}\sum_{k=1}^N\left\|\mathbf{x}(k)-\mathbf{x}^{ref}\right\|_Q^2+\left\|\mathbf{u}(k+1)-\mathbf{u}(k)\right\|_R^2\\
s.t.\begin{cases}\mathbf{x}(k+1)=\mathbf{f}(\mathbf{x}(k),\mathbf{u}(k))+ \mathbf{d}(k),\\
\mathbf{u}_{min}\leq \mathbf{u}(k) \leq \mathbf{u}_{max},\\
\Delta \mathbf{u}_{min}\leq \mathbf{u}(k+1)-\mathbf{u}(k)\leq \Delta \mathbf{u}_{max},
\end{cases}
\label{eq: learning mpc}
\end{aligned}
\end{equation}
where \( Q \) and \( R \) are the weighting matrices, \( \textbf{x} = [V, \beta, r]^T \) represents the state variables, and \( \textbf{u} = [\delta, F_{xr}]^T \) represents the control variables. \( \textbf{f}(\textbf{x}(k), \textbf{u}(k)) \) denotes the vehicle's nominal model, while \( \textbf{d}(k) \) represents the compensation for the vehicle model error derived from \gls{gpr}. The loss function indicates that the vehicle state should track the reference state obtained from the learning-based state planner module. The first term ensures continuous tracking accuracy of the drift state, while the second term ensures that the difference in control input remains within a reasonable range, avoiding excessive abrupt changes.

\section{Simulation Result}
In this section, the proposed algorithm is verified using the Matlab-Carsim platform. This platform combines the analytical system design tools of Matlab with Carsim's realistic vehicle dynamics simulation to accurately represent the vehicle's actual condition. The vehicle model and related parameters used for the experiment are shown in Table \ref{table:simulation parameters}. \par

\subsection{Simulation Set Up}
First, on the given track trajectory, as shown in Fig. \ref{fig:plan_traj}, the vehicle state is planned using the minimum time planner. Based on the planned states, the non-drift state is controlled by the \gls{pp} controller, while the drift state is controlled by the \gls{aedgpr} controller. In the first lap, the traditional \gls{mpc} control results are obtained by running the \gls{mpc} controller without \gls{gpr} while simultaneously collecting \gls{gpr} data. In the second lap, based on the data collected from the traditional \gls{mpc}, a double-layer \gls{gpr} compensation is introduced into the simulation experiments, which are iterated over multiple laps until convergence. Finally, based on the results of the double-layer \gls{gpr} compensation, the speed set \( V_{\text{sets}} \)  for exploration is generated. Then the active exploration algorithm mentioned in the active explorer module is employed until the entire speed sets \( V_{\text{sets}} \) have been explored.\par

\begin{table}[!htbp]
\centering
\vspace{20pt}
\caption{Vehicle and related parameters.}
\begin{tabular}{cccc}
\toprule
Parameters&Value &Parameters &Value \\
\midrule
$m$   &1835$kg$                 &B         &10.92 \\
$I_z$ &3234 $kg \cdot m^2$      &C         &1.458  \\
$a$   &1.4 $m$                  &$P$       &$diag(2,1,1)$  \\
$b$   &1.65 $m$                 & $\lambda$  &0.55 \\
$\mu$       &1.0              & $Q$  &$diag(1, 700, 500)$ \\
$w_1$       &100              & $R$  &$diag(10, 0.001)$ \\
$w_2$       &5  \\ 

\bottomrule
\end{tabular}
\label{table:simulation parameters}
\end{table}

\subsection{Simulation Result} \vspace{-6pt}

\begin{figure}[htbp]
    \centering
    \vspace{20pt}
    \includegraphics[width=0.5\textwidth]{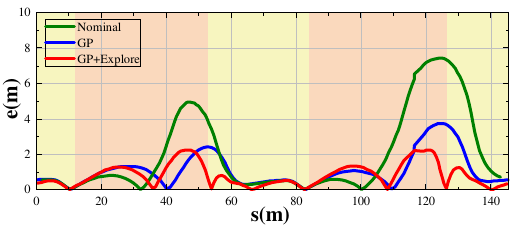}
    \caption{Lateral error in the simulation.}
    \label{fig: Position RMSE Error}
\end{figure}

\begin{figure}[H]
    \centering
    \vspace{8pt}
    \includegraphics[width=0.4\textwidth]{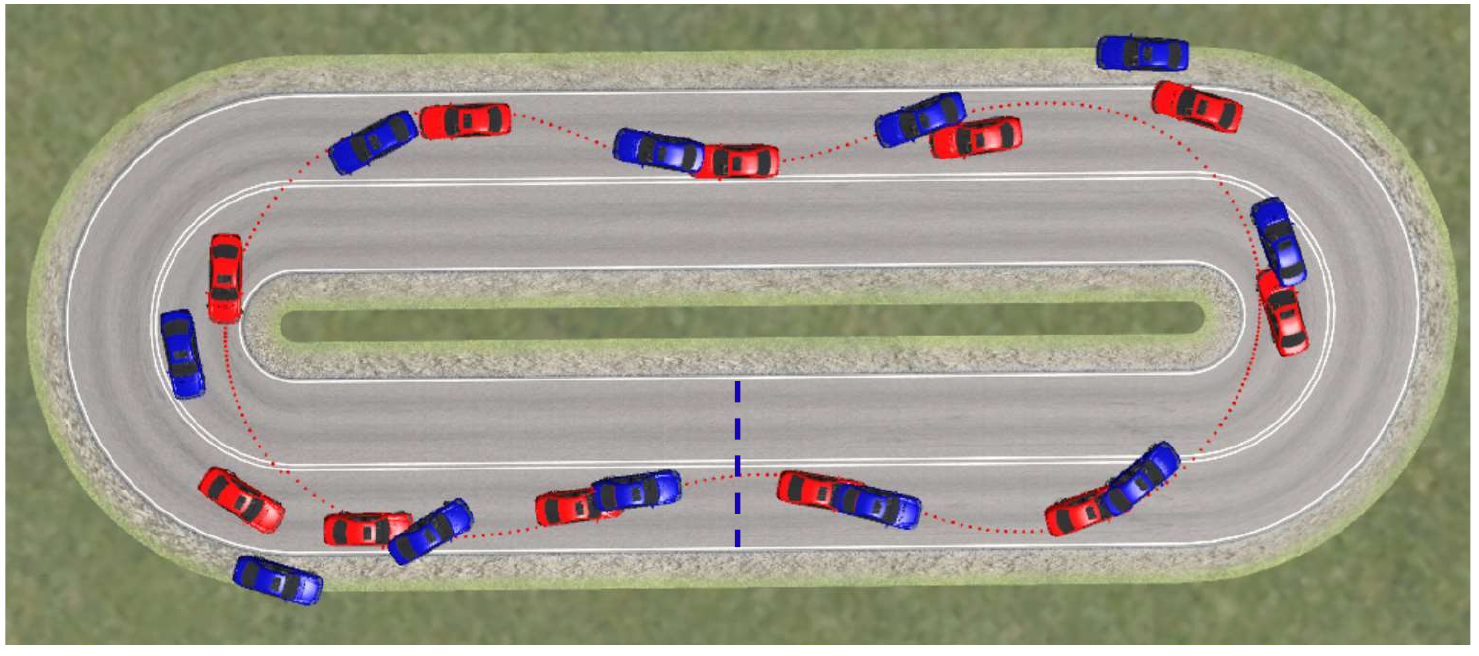}
    \caption{ Comparison of vehicle posture in CarSim. The blue car illustrates the vehicle's posture during cornering, controlled by the \gls{aedgpr} controller without exploration, while the red car shows the posture after exploration. The blue dashed line indicates the vehicle's starting position and the red dashed line represents the reference trajectory.} 
    \label{fig:carsim}
\end{figure}

The comparison of lateral errors across the three experiments is shown in Fig. \ref{fig: Position RMSE Error}. The pink block represents the section controlled by the \gls{aedgpr} controller during drift cornering, while the yellow block corresponds to the section controlled by the \gls{pp} controller. The average lateral error for the \gls{mpc} controller without \gls{gpr} is 2.50 m, while with \gls{gpr} compensation, it is reduced to 1.18 m. After exploration, the average lateral error with \gls{gpr} control further decreases to 0.86 m. The lateral error with \gls{gpr} is 52.8\% lower than the non-\gls{gpr} case, and the error after \gls{gpr} exploration is reduced by 27.1\% compared to the case without exploration. Fig. \ref{fig:carsim} shows a comparison of the vehicle's posture in the simulation. The results demonstrate that the \gls{gpr} compensation model without exploration (blue vehicle) is insufficient for this scenario, highlighting the importance of exploration in \gls{gpr}. The posture of the vehicle with \gls{gpr} exploration (red vehicle) is significantly improved compared to the vehicle without \gls{gpr} exploration.\par

\begin{figure}[h]
    \centering
    \includegraphics[width=0.45\textwidth]{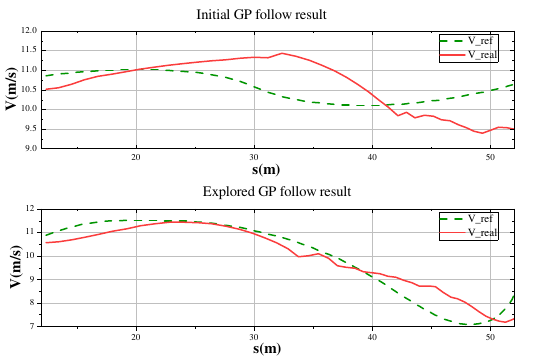}
    \caption{Comparison of speed tracking deviation in the simulation.}
    \label{fig: Velicity tracking results}
\end{figure}

\begin{figure}[h]
   
    \centering
    \includegraphics[width=0.7\linewidth]{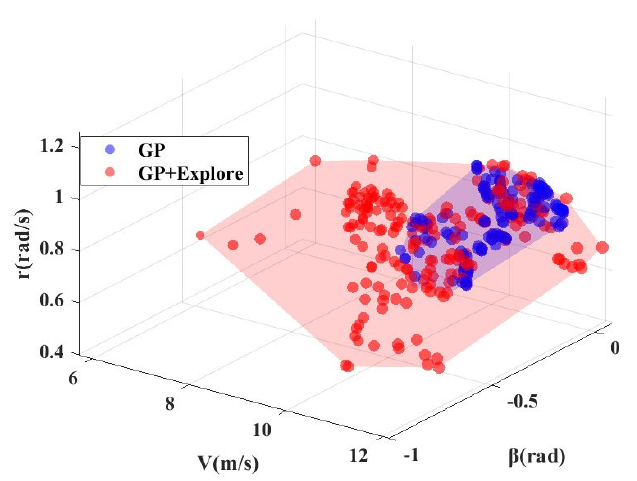}
    \caption{Comparison of the \gls{gpr} dataset distribution in the simulation.}
    \label{fig: discretization}
\end{figure}

Fig. \ref{fig: Velicity tracking results} shows the velocity tracking result between the explored \gls{gpr} and Initial \gls{gpr} without exploration in the cornering. The velocity tracking \gls{rmse} of the \gls{gpr} after exploration is reduced by 10.6\% compared to the \gls{gpr} without exploration. The significant performance improvement after exploration is primarily due to the enrichment of the \gls{gpr} dataset. As shown in Fig. \ref{fig: discretization}, the dataset after exploration covers a broader state space, allowing for a better depiction of the vehicle model, which in turn enhances the control accuracy.\par

\section{Experimental Result}
\begin{figure}
    \vspace{8pt}
    \centering
    \includegraphics[width=1\linewidth]{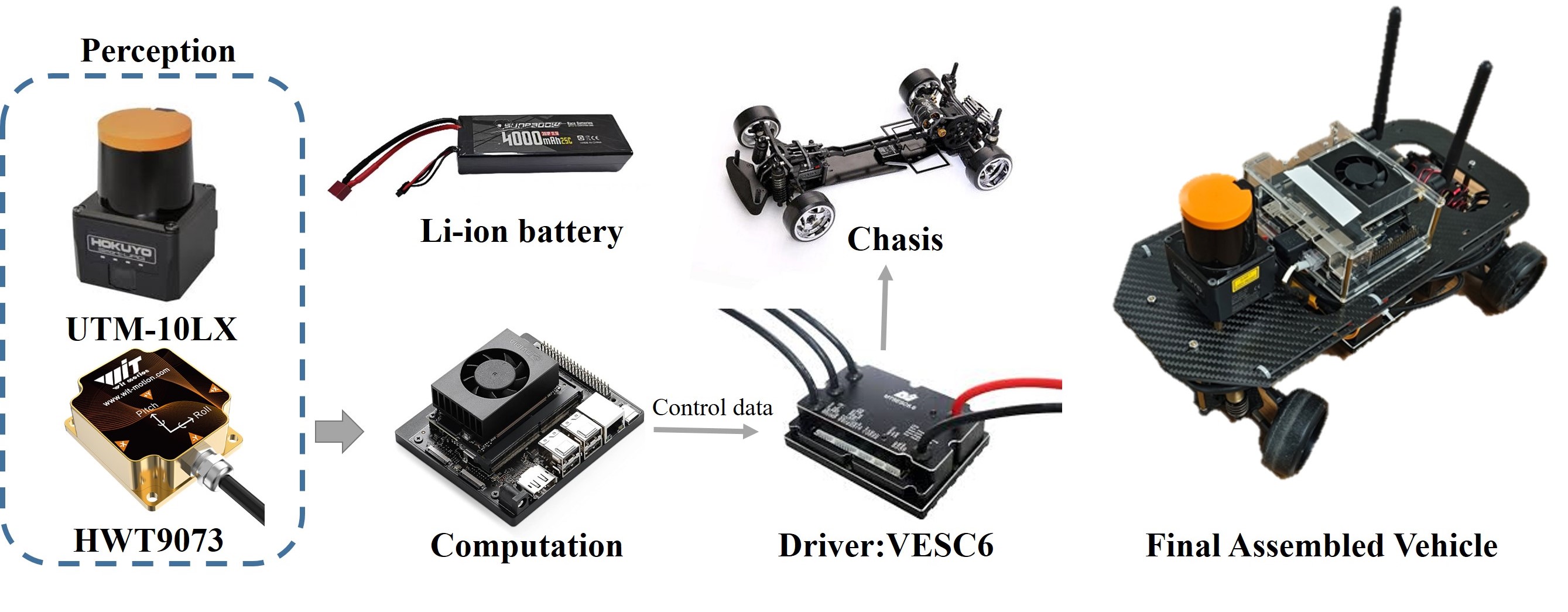}
    \caption{Structural Diagram of the 1:10 scale RC car:
Components for Perception, Computation, Power, and Control.}
    \label{fig:real_car_struct}
\end{figure}

The algorithm is validated on a 1:10 scale RC car. The computational unit is powered by a Jetson Orin NX device running ROS Noetic on Ubuntu 18.04. The final vehicle structure is shown in Fig. \ref{fig:real_car_struct}. Map construction and localization are achieved using an IMU HWT9073, a 2D LiDAR UST-10LX, and odometry data, processed through Cartographer \cite{hess2016real}. The experimental parameters are shown in Table \ref{table:real_car parameters}.\par

\begin{figure}[thpb]
    
    \centering
    \includegraphics[width=0.4\textwidth]{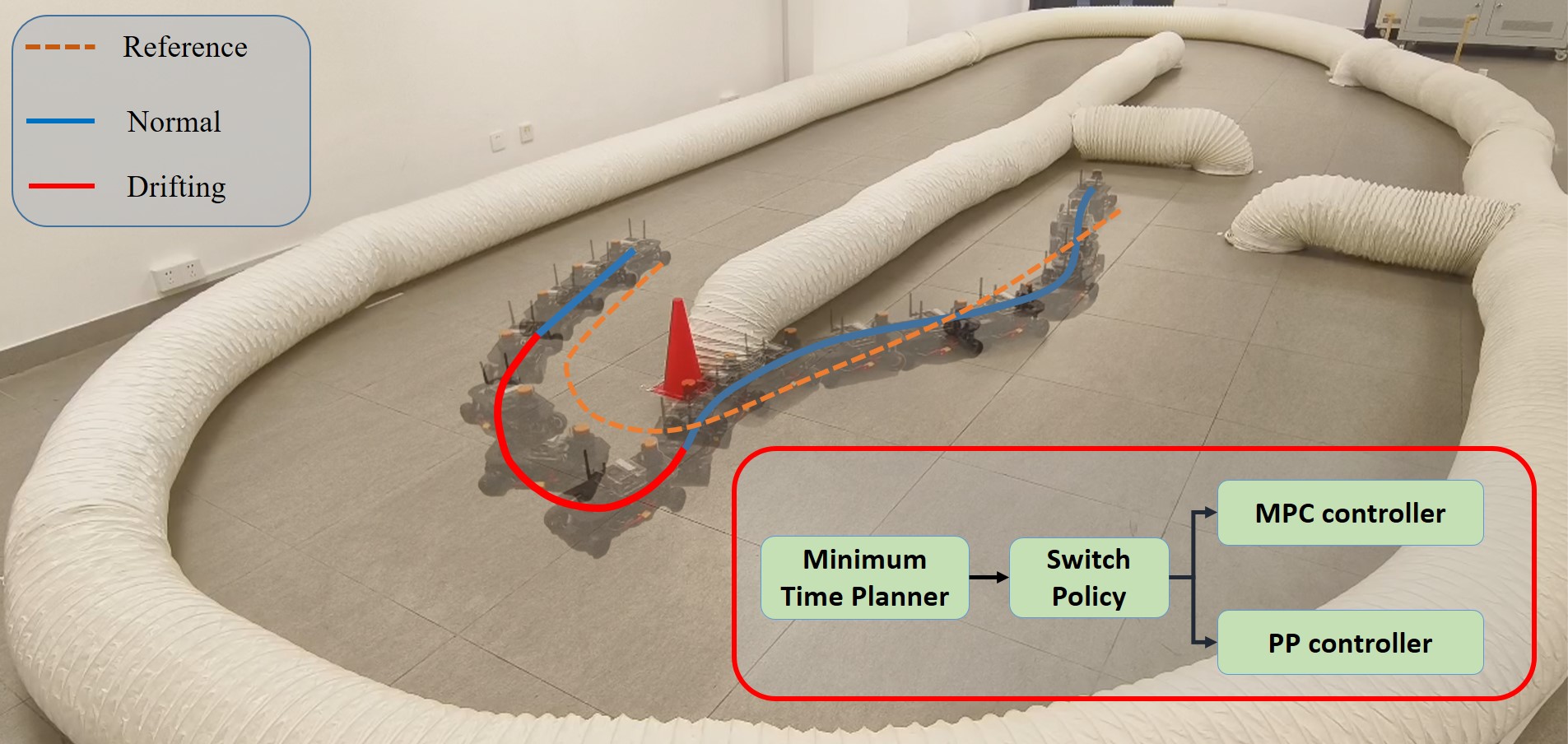}
    \caption{Demonstration of vehicle corner drifting.}
    \label{fig:real_car_global_result}
\end{figure}

The cornering posture of the RC car using the explored \gls{gpr} model is shown in Fig. \ref{fig:real_car_global_result}. We employ both an \gls{mpc} controller and a \gls{pp} controller, switching between them based on the reference state determined by the minimum time trajectory planner. The \gls{aedgpr} controller is utilized for drift cornering, whereas the \gls{pp} controller is engaged for vehicle control during normal driving conditions. The experimental process is similar to the simulation experiments described above.

\begin{table}[h]
\caption{1:10 scale RC car and related parameters.}
\label{table:real_car parameters}
\centering

\begin{tabular}{cccc}
\toprule
Parameters&Value &Parameters &Value \\
\midrule
$m$   &2.356 $kg$                 &B         &18.10\\
$I_z$ &0.0218 $kg \cdot m^2$      &C         &1.323 \\
$a$   &0.122 $m$                  &$P$       &$diag(2,1,1)$  \\
$b$   &0.13 $m$                 & $\lambda$ &0.52 \\
$\mu$       & 0.90          & $Q$  &$diag(10, 600, 500)$\\
$w_1$       &90             & $R$  &$diag(1, 0.001)$ \\
$w_2$       &3  \\ 
\bottomrule
\end{tabular}

\end{table}

\begin{figure}[thpb]
    \centering
    \includegraphics[width=0.45\textwidth]{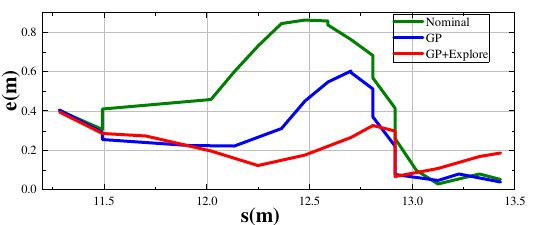}
    \caption{Lateral error in the experiment.}
    \label{fig: Position RMSE Error of real car}
\end{figure}

Fig. \ref{fig: Position RMSE Error of real car} presents the experimental results comparing different algorithms during cornering. The results show that the average lateral error with the nominal model \gls{mpc} control is 0.49 m, while the model with \gls{gpr} compensation reduces this to 0.31 m. The explored \gls{gpr} model further decreases the deviation to 0.22 m. With the inclusion of \gls{gpr}, the average lateral error is reduced by 36.7\% compared to the case without \gls{gpr}, and after exploration, the error decreases by an additional 29.0\%, highlighting the effectiveness of exploration in enhancing accuracy.

\begin{figure}[H]
    \centering
    \includegraphics[width=0.4\textwidth]{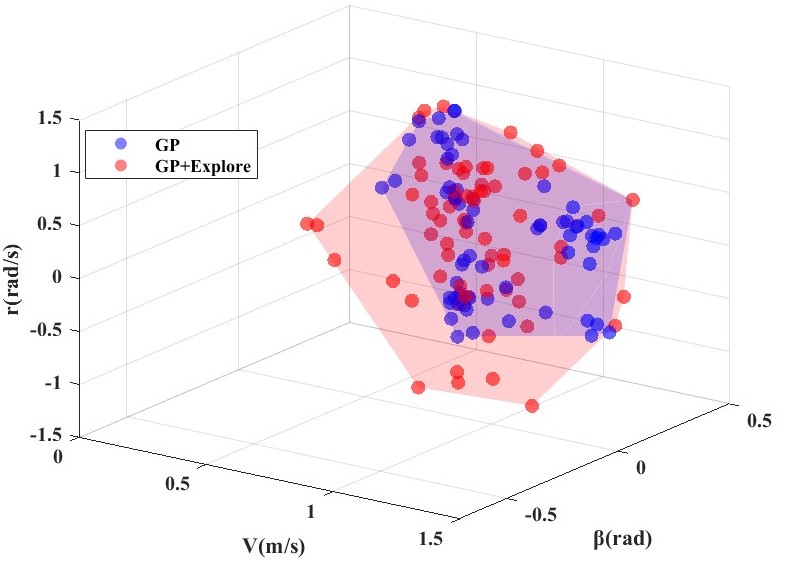}
    \caption{Comparison of \gls{gpr} dataset distributions in the experiment.}
    \label{fig: gp distribution of rc vehicle}
\end{figure}

Additionally, the velocity tracking \gls{rmse} with the \gls{gpr} after exploration is reduced by 7.2\% compared to the \gls{gpr} without exploration. This significant performance improvement is primarily due to the enrichment of the \gls{gpr} dataset. As shown in Fig. \ref{fig: gp distribution of rc vehicle}, the explored \gls{gpr} datasets are more comprehensive and contain fewer redundant points, further improving the model's accuracy.

\section {CONCLUSION}
In this paper, we propose an algorithmic framework, \gls{aedgpr}, which integrates a double-layer \gls{gpr} to correct the vehicle model and actively explores cornering drift states to enrich the \gls{gpr} datasets. The algorithm is validated through both simulations and experiments conducted on a 1:10 scale RC car. In the simulation, the average lateral error with \gls{gpr} is reduced by 52.8\% compared to the case without \gls{gpr}, and after exploration, the error is further decreased by 27.1\%. The velocity tracking \gls{rmse} deviation also decreases by 10.6\% with exploration. In the RC car experiment, the average lateral error with \gls{gpr} is 36.7\% lower, and after exploration, it is reduced by an additional 29.0\%. The velocity tracking \gls{rmse} deviation also improves, showing a reduction of 7.2\% after exploration. \par

In the future, we plan to scale this algorithm to full-sized vehicles and evaluate its performance across a range of extreme conditions.\par
\bibliographystyle{IEEEtran}     
\bibliography{IEEEabrv,ref}

\end{document}